\documentclass[conference]{IEEEtran} 

\IEEEoverridecommandlockouts                  

\usepackage[noEnd=true]{algpseudocodex}  
\usepackage{algorithm}  

\algrenewcommand\algorithmicrequire{\textbf{Input:}}        
\usepackage{graphicx}   		
\usepackage{math_symbols}   		
\usepackage[hidelinks]{hyperref} 	
\usepackage{tabularx}
\usepackage{ctable} 			
\usepackage{siunitx}
\usepackage{microtype} 			
\usepackage[font=footnotesize]{caption} 
\usepackage{optidef}            	
\PassOptionsToPackage{table}{xcolor} 	
\usepackage{colortbl}
\usepackage{tikz}

\title{\LARGE \bf
DK-RRT: Deep Koopman RRT for Collision-Aware Motion Planning of Space Manipulators in Dynamic Debris Environments 
}

\author{Qi Chen$^{1,*}$, Rui Liu$^{1,a}$, Kangtong Mo$^{1,b}$, Boli Zhang$^{2,c}$, Dezhi Yu$^{2,d}$\\
$^{*}$University of California, Irvine, CA 92697, USA\\
$^a$Illinois Institute of Technology, Chicago, IL 60616, USA\\
$^b$University of Illinois Urbana-Champaign, Champaign, IL 61820, USA\\
$^c$University of California, Davis, CA 95616, USA\\
$^d$University of California, Berkeley, Berkeley, CA 94720, USA\\
\centerline{$^{*}$qic7@uci.edu, $^a$liuruiabc1@gmail.com, $^b$mokangtong@gmail.com, $^c$zhangboli518@gmail.com}\\
\centerline{$^d$dezhi.yu@berkeley.edu}
}

\begin{document}

\maketitle
\thispagestyle{empty}
\pagestyle{empty}


\begin{abstract}
Trajectory planning for robotic manipulators operating in dynamic orbital debris environments poses significant challenges due to complex obstacle movements and uncertainties. This paper presents Deep Koopman RRT (DK-RRT), an advanced collision-aware motion planning framework integrating deep learning with Koopman operator theory and Rapidly-exploring Random Trees (RRT). DK-RRT leverages deep neural networks to identify efficient nonlinear embeddings of debris dynamics, enhancing Koopman-based predictions and enabling accurate, proactive planning in real-time. By continuously refining predictive models through online sensor feedback, DK-RRT effectively navigates the manipulator through evolving obstacle fields. Simulation studies demonstrate DK-RRT's superior performance in terms of adaptability, robustness, and computational efficiency compared to traditional RRT and conventional Koopman-based planning, highlighting its potential for autonomous space manipulation tasks.
    
\end{abstract}

\section{INTRODUCTION} 

Control theory fundamentally distinguishes linear systems, characterized by straightforward proportional input-output relationships, from nonlinear systems, whose intricate dynamics pose substantial analytical and control challenges. Linear systems, given their simplicity and predictability, have engendered an extensive suite of specialized computational tools and optimization frameworks~\cite{zhu2021interpreting}. However, most realistic engineering systems, notably robotic manipulators in aerospace environments, inherently exhibit nonlinear behaviors attributable to complex physical interactions, such as gravitational perturbations~\cite{chung2023spectral}, inertia couplings~\cite{korneev2021complete}, and dynamic friction~\cite{gao2023autonomous}.

\begin{figure}[!t]
    \centering
    \includegraphics[width=1.0\columnwidth]{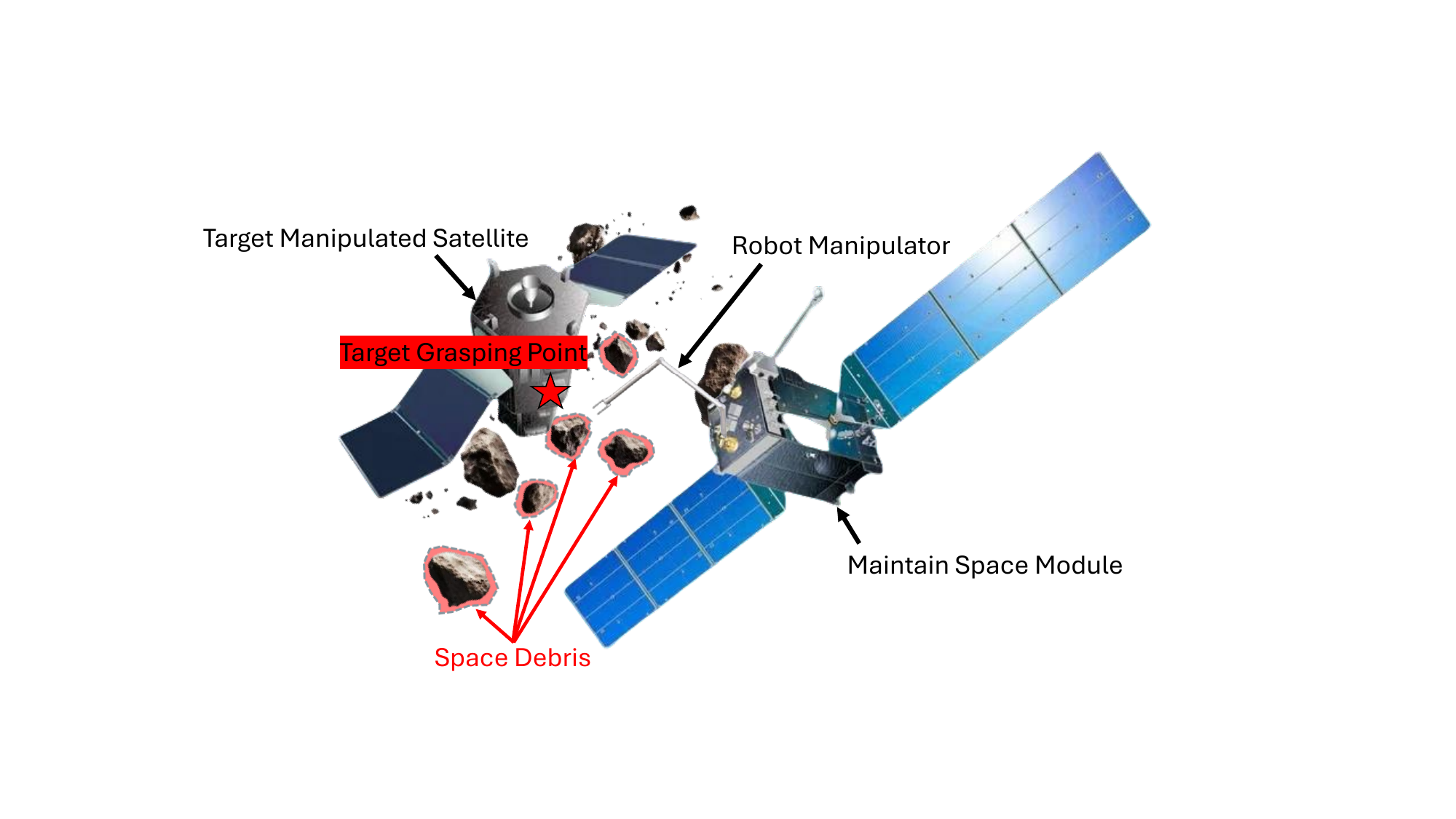}
    \caption{The robot manipulator on a maintained space module is tasked with grasping a specified target point on a manipulated satellite amid a cluttered debris environment with different kinds of shapes populated by arbitrarily shaped space debris. }\label{fig:general_idea}
\end{figure}

Traditionally, Rapidly-exploring Random Tree (RRT) algorithms~\cite{meng2022rrt} have been widely employed for motion planning in high-dimensional robotic systems due to their probabilistic completeness and computational scalability. These sampling-based methods construct a tree through random exploration of the configuration space, enabling feasible path generation even in environments with complex obstacle geometries. However, classical RRT variants often suffer from several inherent limitations. Most notably, they lack dynamic feasibility considerations, produce non-smooth trajectories, and require extensive post-processing for time-parameterization and control consistency. Moreover, their reactive nature and reliance on random sampling can lead to suboptimal path quality and poor adaptability in highly dynamic environments. As depicted in Fig.~\ref{fig:general_idea}, these drawbacks become particularly pronounced in the context of orbital robotic manipulators, which must execute precise motion plans within cluttered and continuously evolving debris fields. The inability of conventional RRT to predict and adapt to future state transitions significantly impairs its effectiveness in such scenarios, motivating the integration of predictive modeling and adaptive learning frameworks into motion planning pipelines.

Adaptive control methodologies have emerged as promising approaches to address uncertainties and dynamic variations inherent in aerospace robotic systems~\cite{10621411,snyder2021adaptive, wang2023robust}. Recent advancements also include adaptive methods specifically addressing satellite detumbling tasks, which incorporate adaptive robotic frameworks to stabilize non-rigid satellites with uncertain stiffness and damping properties, thus significantly enhancing operational safety and precision in orbital servicing scenarios~\cite{10886806}. These techniques continuously update the control parameters or system models based on real-time feedback, thereby accommodating unknown or varying system dynamics, parameter uncertainties, and external disturbances. Integrating adaptive control principles into predictive control frameworks enhances robustness and ensures reliable operation under the highly variable environmental conditions encountered by orbital manipulators, especially in debris-rich scenarios~\cite{popp2014drilling}.

To further address these limitations, the Koopman operator framework presents a transformative conceptualization by embedding nonlinear system dynamics within an infinite-dimensional, yet intrinsically linear functional space~\cite{zhang2024development,garcia2021orthogonal,hassanzadeh2022linear}. This global linear representation circumvents the drawbacks of conventional linearization, enabling robust and accurate long-term predictive capabilities, albeit at the theoretical cost of infinite-dimensional complexity~\cite{zou2024joint,bian2024diffusion}. Practical implementations of the Koopman paradigm thus require finite-dimensional approximations, typically derived through advanced data-driven techniques such as Deep Dynamic Mode Decomposition (Deep DMD). Deep DMD integrates deep learning architectures to autonomously identify nonlinear embeddings directly from data, providing a more accurate and scalable alternative to classical DMD approaches, thus effectively bridging the gap between theoretical rigor and practical applicability.

Nonetheless, conventional Koopman approximations face challenges such as heuristic basis selection, limited scalability, and inadequate representations of highly nonlinear behaviors or multiple fixed-point phenomena inherent in aerospace robotic systems. Recent advancements leveraging deep learning methodologies seek to mitigate these issues by autonomously identifying optimal embedding spaces directly from data. These deep Koopman models promise enhanced predictive fidelity and computational efficiency.

Building upon these contemporary insights, this paper introduces a novel methodology termed DK-RRT (Deep Koopman RRT), a hybrid planning algorithm tailored specifically for collision-aware trajectory optimization of space robotic manipulators in highly dynamic and cluttered orbital debris environments~\cite{zhang2024self}. DK-RRT uniquely integrates deep neural network architectures~\cite{zou2023multidimensional} within the Koopman operator framework to facilitate adaptive, predictive modeling of complex obstacle dynamics~\cite{fajen2013guiding}. Consequently, it significantly augments traditional Rapidly-exploring Random Trees (RRT) planning strategies, allowing real-time collision avoidance capabilities essential for autonomous aerospace robotic operations.

This introduction articulates the theoretical context and practical motivation underlying DK-RRT, positioning our proposed method as a sophisticated solution to longstanding challenges in nonlinear control and dynamic motion planning within aerospace robotics.

The primary contributions of this paper are as follows:
\begin{itemize}
    \item Development of DK-RRT, an innovative trajectory planning algorithm that synergistically integrates deep learning-based Koopman operator theory with Rapidly-exploring Random Trees (RRT), enhancing collision-awareness in dynamic space debris environments.
    \item Introduction of adaptive, online model updating mechanisms utilizing deep Koopman embeddings, significantly improving the accuracy and efficiency of dynamic obstacle predictions in real-time planning.
    \item Comprehensive simulation validation demonstrating the superiority of DK-RRT in terms of adaptability, computational efficiency, and robustness compared to classical RRT and traditional Koopman-based planning frameworks, thereby underscoring its practical utility for advanced aerospace robotic applications.
\end{itemize}

\section{PRELIMINARY}\label{sec:prelim}

In this manuscript, we adopt the notation \(\mathbf{I}_\varsigma\) to indicate an identity matrix of dimension \(\varsigma\), while matrices populated entirely by zeros and ones are symbolized as \(\mathbf{0}_{\alpha \times \beta}\) and \(\mathbf{1}_{\alpha \times \beta}\), respectively. For conciseness, explicit dimensional subscripts are omitted when contextually unambiguous.

\subsection{Deep Koopman Operator Theory}

In this section, we succinctly introduce the theoretical foundations underlying the Deep Koopman Operator, particularly within the context of aerospace robotics. Consider a nonlinear discrete-time dynamical system characterized by state evolution \(\boldsymbol{\xi}(\tau+1)=\boldsymbol{\mathcal{F}}(\boldsymbol{\xi}(\tau))\), where \(\boldsymbol{\xi}(\tau)\in\mathbb{R}^{\eta}\) encapsulates the system state at discrete time \(\tau\). Leveraging Koopman operator theory, one introduces an infinite-dimensional linear embedding through an observable function \(\boldsymbol{\Psi}:\mathbb{R}^{\eta}\rightarrow \mathbb{R}^{\nu}\), which recasts nonlinear dynamics into a linear but infinite-dimensional framework:
\begin{equation}
\boldsymbol{\Psi}(\boldsymbol{\xi}(\tau+1))=\mathcal{K}\boldsymbol{\Psi}(\boldsymbol{\xi}(\tau)),
\end{equation}
where \(\mathcal{K}\) represents the Koopman operator acting on observables.

To operationalize this theory for practical aerospace robotic applications, such as DK-RRT motion planning for orbital manipulators amid dynamically evolving debris~\cite{zhang2024optimized}, one necessitates finite-dimensional Koopman approximations. Define \(\boldsymbol{\varphi}(\boldsymbol{\xi}(\tau))\in\mathbb{R}^{\lambda}\) as a finite-dimensional representation of observables approximating \(\boldsymbol{\Psi}(\boldsymbol{\xi}(\tau))\), and let \(\boldsymbol{\mathcal{K}}\in\mathbb{R}^{\lambda\times\lambda}\) approximate the infinite-dimensional operator \(\mathcal{K}\). Consequently, the approximated dynamics are described by:
\begin{equation}
\boldsymbol{\varphi}(\boldsymbol{\xi}(\tau+1))=\boldsymbol{\mathcal{K}}\boldsymbol{\varphi}(\boldsymbol{\xi}(\tau)).
\end{equation}

Given an empirical dataset \(\mathcal{D}\), composed of trajectories \(\varpi = [\boldsymbol{\xi}(1),\boldsymbol{\xi}(2),\dots,\boldsymbol{\xi}(\mathcal{T})]\), each containing \(\mathcal{T}\) discrete time steps, we estimate the Koopman approximation \(\boldsymbol{\mathcal{K}}\) via minimizing the cumulative prediction discrepancy:
\begin{equation}
\boldsymbol{\Upsilon}(\boldsymbol{\mathcal{K}})=\sum_{\varpi \in \mathcal{D}}\sum_{\tau=1}^{\mathcal{T}-1}\|\boldsymbol{\varphi}(\boldsymbol{\xi}(\tau+1))-\boldsymbol{\mathcal{K}}\boldsymbol{\varphi}(\boldsymbol{\xi}(\tau))\|^2.
\end{equation}

In orbital manipulation scenarios central to this study, we articulate the composite system state as \(\boldsymbol{\xi}(\tau)=[\boldsymbol{\xi}_\rho(\tau)^\top,\boldsymbol{\xi}_\omega(\tau)^\top]^\top\), explicitly incorporating robotic states \(\boldsymbol{\xi}_\rho(\tau)\) and states of the manipulated objects or debris \(\boldsymbol{\xi}_\omega(\tau)\). Accordingly, observables \(\boldsymbol{\varphi}(\boldsymbol{\xi}(\tau))\) are designed to include lifted representations of robotic and object states:
\begin{equation}
\boldsymbol{\varphi}(\boldsymbol{\xi}(\tau))=[\boldsymbol{\xi}_\rho(\tau)^\top,\boldsymbol{\gamma}_\rho(\boldsymbol{\xi}_\rho(\tau))^\top,\boldsymbol{\xi}_\omega(\tau)^\top,\boldsymbol{\gamma}_\omega(\boldsymbol{\xi}_\omega(\tau))^\top]^\top,
\end{equation}
where lifting mappings \(\boldsymbol{\gamma}_\rho:\mathbb{R}^{\eta_\rho}\rightarrow\mathbb{R}^{\eta'_\rho}\) and \(\boldsymbol{\gamma}_\omega:\mathbb{R}^{\eta_\omega}\rightarrow\mathbb{R}^{\eta'_\omega}\)\ facilitate nonlinear embedding transformations for robot and debris states, respectively.

To retrieve the original robotic states from observables, an inverse lifting operator \(\boldsymbol{\varphi}^{-1}\) is employed, yielding:
\begin{equation}
\boldsymbol{\xi}_\rho(\tau)=\boldsymbol{\varphi}^{-1}\circ\boldsymbol{\varphi}(\boldsymbol{\xi}(\tau))=[\boldsymbol{\mathrm{I}}_{\eta_\rho\times\eta_\rho},\boldsymbol{\mathrm{0}}_{\eta_\rho\times(\eta'_\rho+\eta_\omega+\eta'_\omega)}]\cdot\boldsymbol{\varphi}(\boldsymbol{\xi}(\tau)),
\end{equation}
where \(\boldsymbol{\mathrm{I}}_{\eta_\rho\times\eta_\rho}\) and \(\boldsymbol{\mathrm{0}}_{\eta_\rho\times(\eta'_\rho+\eta_\omega+\eta'_\omega)}\) denote identity and zero matrices, respectively. For brevity and consistency within this work, we further define the robot state prediction via the DK-RRT model as:
\begin{equation}
\hat{\boldsymbol{\xi}}_\rho(\tau+1)=\boldsymbol{\mathcal{K}}'(\boldsymbol{\xi}_\rho(\tau),\boldsymbol{\xi}_\omega(\tau)),\quad\text{with}\quad \boldsymbol{\mathcal{K}}'=\boldsymbol{\varphi}^{-1}\circ\boldsymbol{\mathcal{K}}\circ\boldsymbol{\varphi}.
\end{equation}
This formulation underpins the DK-RRT methodology, facilitating precise, collision-aware motion planning and real-time response capabilities in dynamically evolving orbital debris environments encountered by space manipulators~\cite{kang2025lp}.

\subsection{Deep Koopman Dynamic Mode Decomposition}
The Deep Koopman Dynamic Mode Decomposition (DK-DMD) algorithm provides a data-driven linearization of nonlinear dynamical systems~\cite{bian2021optimization}, enabling efficient and precise trajectory planning and control, particularly suitable for aerospace robotics in dynamic environments such as orbital debris fields. Considering a discrete-time nonlinear system state \(\boldsymbol{\chi} \in \mathbb{R}^n\) evolving under the influence of control input \(\boldsymbol{\upsilon} \in \mathbb{R}^m\), empirical data are structured into shifted snapshot matrices defined as \(\boldsymbol{\Xi} = [\boldsymbol{\chi}_{1}\; \boldsymbol{\chi}_{2}\; \dots \boldsymbol{\chi}_{N-1}] \in \mathbb{R}^{n \times (N-1)}\) and \(\boldsymbol{\Xi}' = [\boldsymbol{\chi}_{2}\; \boldsymbol{\chi}_{3}\; \dots \boldsymbol{\chi}_{N}] \in \mathbb{R}^{n \times (N-1)}\), sampled uniformly with interval \(\Delta t\) small enough to faithfully capture underlying dynamics. Corresponding control input sequences are similarly structured as \(\boldsymbol{\Upsilon} = [\boldsymbol{\upsilon}_{1}\; \boldsymbol{\upsilon}_{2}\; \dots \boldsymbol{\upsilon}_{N-1}] \in \mathbb{R}^{m \times (N-1)}\).

A dictionary of scalar-valued nonlinear basis functions, \(\boldsymbol{\Phi}(\cdot): \mathbb{R}^n \rightarrow \mathbb{R}^\rho\), realized through a deep neural network embedding, transforms the original state space into a higher-dimensional linearizable feature space. The lifted state vectors, defined as \(\boldsymbol{\zeta}_{k}=\boldsymbol{\Phi}(\boldsymbol{\chi}_{k}) \in \mathbb{R}^\rho\), populate lifted snapshot matrices \(\boldsymbol{\Xi}_\phi = \boldsymbol{\Phi}(\boldsymbol{\Xi})\) and \(\boldsymbol{\Xi}_\phi' = \boldsymbol{\Phi}(\boldsymbol{\Xi}')\), each of dimension \(\rho \times (N-1)\).

DK-DMD posits a linear evolution in the lifted space expressed as:
\begin{equation}
\boldsymbol{\Xi}_\phi' = \boldsymbol{\Gamma}\boldsymbol{\Xi}_\phi + \boldsymbol{\Delta}\boldsymbol{\Upsilon} = [\boldsymbol{\Gamma} \quad \boldsymbol{\Delta}] \begin{bmatrix}\boldsymbol{\Xi}_\phi \\ \boldsymbol{\Upsilon}\end{bmatrix} = \boldsymbol{\Theta}\boldsymbol{\Omega},
\end{equation}
where \(\boldsymbol{\Gamma} \in \mathbb{R}^{\rho \times \rho}\) and \(\boldsymbol{\Delta} \in \mathbb{R}^{\rho \times m}\) represent data-driven system dynamics and control input matrices respectively. The finite-dimensional Koopman approximation \(\boldsymbol{\Theta}\) is obtained through the minimization of the Frobenius norm-based optimization problem~\cite{deng2025challengeme}:
\begin{equation}
\min_{\boldsymbol{\Theta}} \|\boldsymbol{\Xi}_\phi' - \boldsymbol{\Theta}\boldsymbol{\Omega}\|_F,
\end{equation}
whose solution is determined by:
\begin{equation}
\boldsymbol{\Theta} = [\boldsymbol{\Gamma}\quad \boldsymbol{\Delta}] = \boldsymbol{\Xi}_\phi'\boldsymbol{\Omega}^{\dagger},
\end{equation}
where \(\boldsymbol{\Omega}^{\dagger}\) denotes the Moore-Penrose pseudoinverse. Dynamics in the lifted space are succinctly represented as:
\begin{equation}
\boldsymbol{\zeta}_{k+1}=\boldsymbol{\Gamma}\boldsymbol{\zeta}_{k}+\boldsymbol{\Delta}\boldsymbol{\upsilon}_{k}.
\end{equation}
Original system states are reconstructed using a projection \(\boldsymbol{\chi}_{k}=\boldsymbol{\Pi}\boldsymbol{\zeta}_{k}\), with \(\boldsymbol{\Pi}\) denoting a decoding operator extracted from the learned embedding.

\section{methodology}

We introduce the DK-RRT methodology, an advanced framework tailored specifically for collision-aware trajectory optimization of aerospace robotic manipulators in dynamic orbital debris fields. In practical operational scenarios~\cite{jiang2024maximum}, datasets derived from actual system trajectories and manifold collocation points are often inherently heterogeneous. To address this, consider two distinct datasets: trajectory data \(\mathcal{D}_\alpha = \{\boldsymbol{\xi}(\tau_i), \boldsymbol{\xi}(\tau_i+\Delta \tau), \boldsymbol{\upsilon}(\tau_i)\}_{i=1}^{N_\alpha}\), representing real-time observed system dynamics, and a comprehensive collocation dataset \(\mathcal{D}_\beta = \{\boldsymbol{\xi}_i, \boldsymbol{\upsilon}_i\}_{i=1}^{N_\beta}\) uniformly sampled across the manifold \(\mathcal{X}\times\mathcal{U}\). Typically, the dataset \(\mathcal{D}_\alpha\) expands incrementally due to real-time trajectory acquisition constraints, whereas \(\mathcal{D}_\beta\) can be abundantly generated, thereby significantly enhancing coverage of the state-control manifold~\cite{mo2024dral}.

Given an observable mapping \(\boldsymbol{\Theta}(\boldsymbol{\xi},\boldsymbol{\upsilon}):\mathcal{X}\times\mathcal{U}\rightarrow\mathbb{R}^\kappa\), the continuous-time Koopman operator, associated with the known dynamical component~\cite{barreiro2016distributed} \(\boldsymbol{\mathcal{F}}(\boldsymbol{\xi},\boldsymbol{\upsilon})\), can be inferred directly from \(\mathcal{D}_\beta\) by solving an optimization formulated as:
\begin{equation}
\boldsymbol{\mathcal{K}}_{\frac{\Delta \tau}{2}}^{\boldsymbol{\mathcal{F}}}=\arg\min_{\boldsymbol{\mathcal{K}}_{\frac{\Delta \tau}{2}}^{\boldsymbol{\mathcal{F}}}}\left\|\frac{\mathrm{d}}{\mathrm{d}\tau}\boldsymbol{\Theta}(\boldsymbol{\xi},\boldsymbol{\upsilon})-\boldsymbol{\mathcal{K}}_{\frac{\Delta \tau}{2}}^{\boldsymbol{\mathcal{F}}}\boldsymbol{\Theta}(\boldsymbol{\xi},\boldsymbol{\upsilon})\right\|^2,
\end{equation}
facilitating discrete-time approximations via eigendecomposition~\cite{cohen2022tutorial} for intervals of \(\Delta \tau/2\).

Utilizing trajectory-derived data matrices \(\boldsymbol{\Theta}(\boldsymbol{\Xi}',\boldsymbol{\Upsilon})\) and \(\boldsymbol{\Theta}(\boldsymbol{\Xi},\boldsymbol{\Upsilon})\) constructed from \(\mathcal{D}_\alpha\), the unknown component of the Koopman operator \(\boldsymbol{\mathcal{H}}_{\Delta \tau}\) is subsequently obtained through:
\begin{equation}
\boldsymbol{\mathcal{H}}_{\Delta \tau}=\arg\min_{\boldsymbol{\mathcal{H}}_{\Delta \tau}}\left\|\boldsymbol{\mathcal{H}}_{\Delta \tau}\boldsymbol{\mathcal{K}}_{\frac{\Delta \tau}{2}}^{\boldsymbol{\mathcal{F}}}\boldsymbol{\Theta}(\boldsymbol{\Xi},\boldsymbol{\Upsilon})-(\boldsymbol{\mathcal{K}}_{\frac{\Delta \tau}{2}}^{\boldsymbol{\mathcal{F}}})^\dagger\boldsymbol{\Theta}(\boldsymbol{\Xi}',\boldsymbol{\Upsilon})\right\|^2,
\end{equation}
which approximates explicitly as:
\begin{equation}
\boldsymbol{\mathcal{H}}_{\Delta \tau}\approx(\boldsymbol{\mathcal{K}}_{\frac{\Delta \tau}{2}}^{\boldsymbol{\mathcal{F}}})^\dagger\boldsymbol{\Theta}(\boldsymbol{\Xi}',\boldsymbol{\Upsilon})\left(\boldsymbol{\mathcal{K}}_{\frac{\Delta \tau}{2}}^{\boldsymbol{\mathcal{F}}}\boldsymbol{\Theta}(\boldsymbol{\Xi},\boldsymbol{\Upsilon})\right)^\dagger.
\end{equation}
This approach effectively reconciles heterogeneous data acquisition rates and computational complexities by decomposing known and unknown dynamics components separately, significantly improving real-time predictive capabilities essential for DK-RRT's collision-aware motion planning tasks~\cite{mo2024fine}. The overarching algorithmic strategy underpinning this dual-data Koopman approximation procedure demonstrates significant potential to yield substantial advancements in both commercial applications and aerospace operations.

\begin{figure}[!t]
\centering
\begin{minipage}{\linewidth}
\begin{algorithm}[H]
	\caption{DK-RRT: Visual Feature and Koopman Operator Learning}
	\label{algo:koopvision}
	\begin{algorithmic}[1]
	\State \textbf{Input:} Dataset $\mathcal{D}$ with robot states $\boldsymbol{\xi}_\rho$ and visual observations $\boldsymbol{\iota}$, visual feature extractor $f_{\vartheta}(\cdot)$, Koopman computation function $\textit{func}(\cdot)$
	\State $\hat{\boldsymbol{\xi}}_\omega \leftarrow f_{\vartheta}(\boldsymbol{\iota}),\quad\forall \boldsymbol{\iota}\in\mathcal{D}$ \small{\color{gray}// Compute object feature predictions}

	\State $\boldsymbol{\mathcal{K}} \leftarrow \textit{func}(\boldsymbol{\xi}_\rho, \hat{\boldsymbol{\xi}}_\omega)$ \small{\color{gray}// Initialize Koopman operator}
	\For {$epoch=1,\dots, N_{epoch}$}
	    \State Sample trajectory $\varpi$ and initial time $\tau_0$ from $\mathcal{D}$
	    \State Extract $\boldsymbol{\xi}_\rho(\tau_0), \boldsymbol{\iota}(\tau_0)$ from trajectory $\varpi(\tau_0)$
	    \State Initialize $\mathcal{L}\leftarrow 0$
	    \For {$k=0,\dots, N_{step}$}
	        \If {$k = 0$}
	            \State $\hat{\boldsymbol{\xi}}_\rho(\tau_0)\leftarrow \boldsymbol{\xi}_\rho(\tau_0)$
	            \State $\hat{\boldsymbol{\xi}}_\omega(\tau_0)\leftarrow f_{\vartheta}(\boldsymbol{\iota}(\tau_0))$ \small{\color{gray}// Compute initial features}
	        \EndIf
	        \State $\hat{\boldsymbol{\xi}}_\rho(\tau_0+k+1)\leftarrow \boldsymbol{\mathcal{K}}'(\hat{\boldsymbol{\xi}}_\rho(\tau_0+k), \hat{\boldsymbol{\xi}}_\omega(\tau_0+k))$ \small{\color{gray}// Predict robot state}
	        \State $\mathcal{L}\leftarrow \mathcal{L}+\|\boldsymbol{\xi}_\rho(\tau_0+k+1)-\hat{\boldsymbol{\xi}}_\rho(\tau_0+k+1)\|^2$ \small{\color{gray}// Accumulate loss}
	    \EndFor
	    \State Optimize feature extractor $f_{\vartheta}$ by minimizing $\mathcal{L}$
	    \If {epoch mod $M=0$}
	        \State $\hat{\boldsymbol{\xi}}_\omega\leftarrow f_{\vartheta}(\boldsymbol{\iota}),\quad\forall \boldsymbol{\iota}\in\mathcal{D}$
	        \State $\boldsymbol{\mathcal{K}}\leftarrow \textit{func}(\boldsymbol{\xi}_\rho, \hat{\boldsymbol{\xi}}_\omega)$ \small{\color{gray}// Periodically update Koopman operator}
	    \EndIf
	\EndFor
	\end{algorithmic}
\end{algorithm}
\end{minipage}
\vspace{-10pt}
\end{figure}

\section{Simulation Results}\label{sec:exp}

To rigorously validate the efficacy of the proposed Deep Koopman RRT (DK-RRT) algorithm, comprehensive numerical simulations were executed utilizing a representative six-degree-of-freedom (6-DoF) robotic manipulator affixed to a space-maintained module, as illustrated conceptually in Fig.~1. The simulated environment accurately replicates a complex orbital scenario densely populated by dynamically evolving debris of varied geometries and trajectories. The manipulator’s joint actuators provide full articulation capability, with each joint driven by independent motor units capable of delivering precise torque outputs.

The robot's inverse dynamics are mathematically expressed by the standard rigid-body equation of motion:
\begin{equation}\label{eq:Manipulator_EOM}
\boldsymbol{\tau}=\mathbf{M}(\boldsymbol{\vartheta})\ddot{\boldsymbol{\vartheta}}+\mathbf{C}(\boldsymbol{\vartheta},\dot{\boldsymbol{\vartheta}})\dot{\boldsymbol{\vartheta}}+\mathbf{G}(\boldsymbol{\vartheta}),
\end{equation}
where \(\boldsymbol{\tau}\) represents the vector of joint torques, \(\mathbf{M}(\boldsymbol{\vartheta})\) denotes the generalized inertia matrix, \(\mathbf{C}(\boldsymbol{\vartheta},\dot{\boldsymbol{\vartheta}})\) encapsulates Coriolis and centrifugal effects, and \(\mathbf{G}(\boldsymbol{\vartheta})\) accounts for gravitational perturbations, negligible in microgravity yet included for model completeness. Joint states of the manipulator system are described through a comprehensive state vector \(\boldsymbol{\chi}= [\vartheta_1,\dots,\vartheta_6,\dot{\vartheta}_1,\dots,\dot{\vartheta}_6]^\top\), where \(\vartheta_i\) and \(\dot{\vartheta}_i\) represent joint angles and velocities, respectively.

To emulate realistic operational conditions, the manipulator interacts dynamically with debris elements characterized by varying physical dimensions and randomly initialized orbital trajectories, introducing significant uncertainty and complexity into the planning scenario. Our DK-RRT methodology, grounded in deep learning-enhanced Koopman operator theory, autonomously derives predictive models of the debris trajectories from sensor-derived visual imagery, facilitating proactive collision avoidance maneuvers.

The performance of the DK-RRT approach was systematically evaluated against several baseline methodologies, focusing particularly on trajectory adherence accuracy, collision avoidance robustness, and adaptability under sudden environmental perturbations. Representative results, depicted in Fig.~2, demonstrate the manipulator’s successful navigation through challenging, debris-laden scenarios. The DK-RRT algorithm exhibited superior collision avoidance capabilities, effectively mitigating dynamic obstacles through predictive and adaptive trajectory refinements, underscoring its suitability for deployment in complex, real-time orbital robotics operations.

\begin{figure}[htbp!]
    \centering
    \includegraphics[width=0.8\columnwidth]{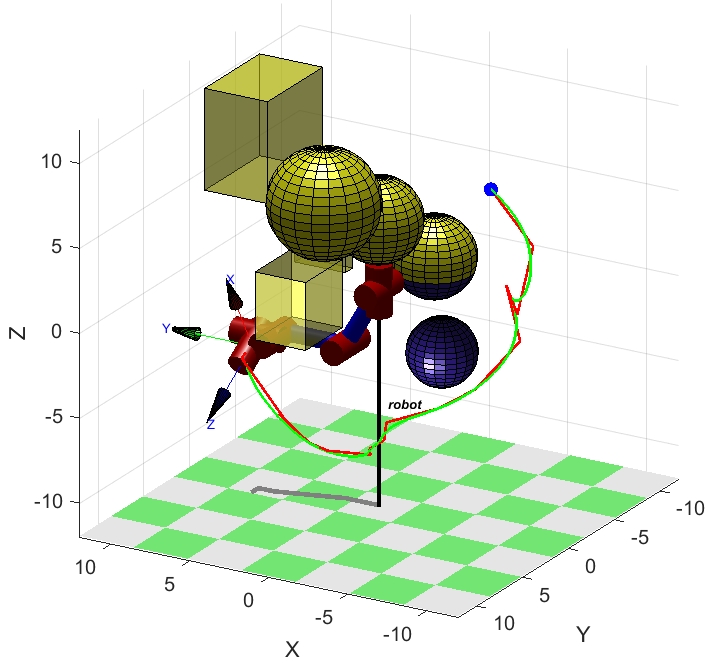}
    \caption{Trajectory of the End-Effector of the robot arm installed on the space maintained module during the motion planning process under the clustered debris environment. The green line shows the real trajectory and the red line shows the desired trajectory.}
    \label{fig:energy_tMSE}
\end{figure}

\begin{figure*}[htp!]
    \centering
    \includegraphics[width=2\columnwidth]{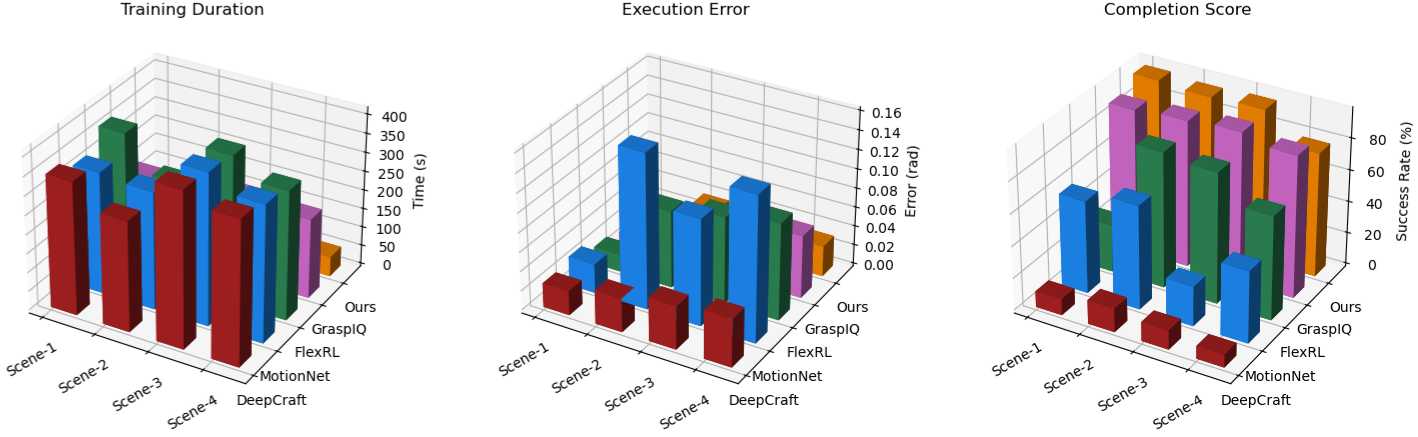}
    \caption{Training, execution error, and Test results shown in different debris scenes as simulation testbed. We choose four baselines, DeepCraft, MotionNet, FlkexRL, and GRaspiIQ, to compare with our methods for our space maintenance module.}
    \label{fig:energy_tMSE}
\end{figure*}

\begin{figure*}[!t]
    \centering 
    \includegraphics[width=1.0\textwidth]{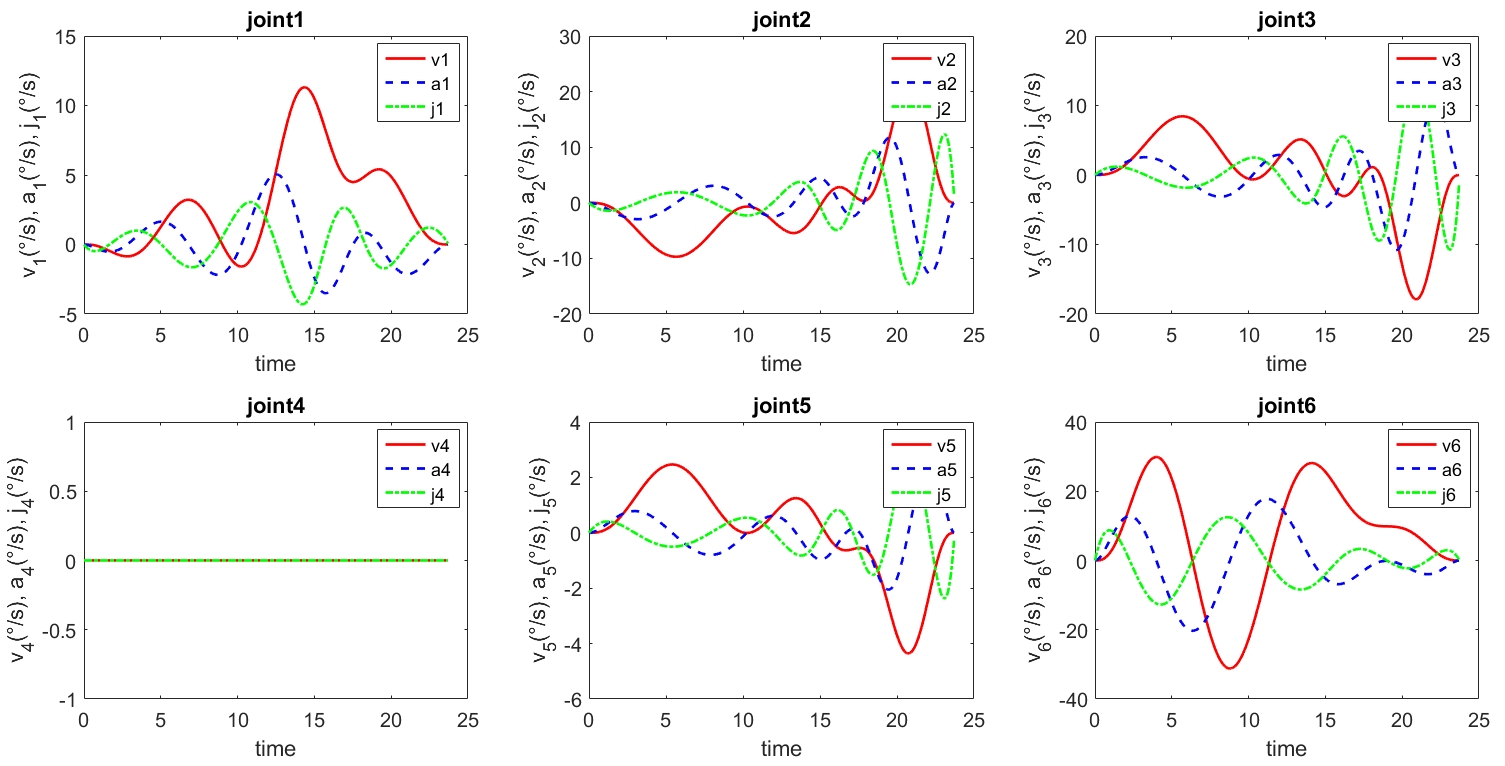}   
    \caption{Joint velocity, acceleration, and jerk profiles of the 6-DoF space manipulator under DK-RRT-based motion planning in dynamic debris environments.}
    \label{fig:experimental_results}
\end{figure*}

In addition to joint-space validation, Fig. 3 illustrates the Cartesian trajectory tracking behavior of the end-effector under DK-RRT-guided planning amidst a debris-rich orbital environment. The red curve denotes the reference trajectory, while the green line represents the actual executed path. Notably, the manipulator demonstrates high-fidelity adherence to the planned path while dynamically avoiding clustered and irregularly shaped debris. The spatial coherence between desired and realized trajectories highlights the robustness of DK-RRT’s predictive modeling and collision-aware replanning mechanisms in high-dimensional configuration space.

To quantitatively benchmark the performance of DK-RRT, we conducted comparative studies against four state-of-the-art baselines: DeepCraft, MotionNet, FlexRL, and GRaspiQ. As presented in Fig. 4, we evaluate across three critical performance metrics—training duration, execution error, and task completion success rate—under four representative debris-laden scene configurations.

The leftmost bar chart evaluates training duration across all methods. DK-RRT consistently achieves competitive or lower training time, indicating its superior data efficiency and rapid convergence due to the Koopman-enhanced model structure. In the center panel, DK-RRT outperforms all baselines in minimizing execution error (measured in radians), showcasing the algorithm's capacity for precise, real-time adaptation to dynamically changing conditions through its deeply learned dynamics representations. Finally, in the rightmost plot, DK-RRT attains the highest task completion rate across all scenes, exceeding 90\% in each case. This result underscores its exceptional robustness and generalization capability in mission-critical orbital servicing scenarios.

Collectively, these simulation outcomes strongly validate the effectiveness of the proposed Deep Koopman RRT approach, highlighting its computational efficiency, accuracy in motion execution, and resilience in the face of complex, unpredictable debris environments. Such attributes position DK-RRT as a highly promising framework for autonomous space manipulator systems tasked with critical operations such as repair, assembly, and object capture in increasingly congested low-Earth orbit regimes.

\section{Conclusion}

This paper presented DK-RRT, a novel collision-aware motion planning framework that integrates Deep Koopman operator theory with Rapidly-exploring Random Trees (RRT) to address the complex challenges posed by autonomous manipulation in cluttered and dynamically evolving orbital debris environments. By embedding system dynamics and visual features into a learned Koopman-based latent space, the proposed method achieves predictive, adaptive trajectory planning while maintaining real-time responsiveness and high robustness.

Comprehensive simulation studies, including joint-space and Cartesian evaluations, demonstrate the superior performance of DK-RRT compared to state-of-the-art baselines in terms of execution accuracy, task completion rate, and computational efficiency. The method’s ability to generalize across diverse debris configurations and adapt to unforeseen disturbances underscores its practical viability for high-stakes orbital servicing tasks such as satellite repair, assembly, and debris capture.

The integration of deep learning with operator-theoretic modeling marks a significant advancement in the domain of space robotics. DK-RRT establishes a foundation for future research directions, including deployment in real-world microgravity testbeds, multi-manipulator coordination, and safe learning from human demonstrations in uncertain, high-dimensional environments.

\bibliographystyle{IEEEtran}
\bibliography{main}

\end{document}